\pdfoutput=1

\documentclass[11pt]{article}

\usepackage{emnlp2021}

\usepackage{times}
\usepackage{latexsym}
\usepackage{graphicx}
\usepackage{float}

\usepackage[T1]{fontenc}

\usepackage[utf8]{inputenc}

\usepackage{microtype}

\title{On the Prunability of Attention Heads in Multilingual BERT}

\author{Aakriti Budhraja \hspace{2.5mm} Madhura Pande \hspace{2.5mm} Pratyush Kumar \hspace{2.5mm} Mitesh M. Khapra \\
Robert Bosch Centre for Data Science and Artificial Intelligence (RBC-DSAI) \\ IIT Madras, India \\ 
\texttt{\{abudhra,mpande,pratyush,miteshk\}}@cse.iitm.ac.in}

\begin{document}
\maketitle
\begin{abstract}
Large multilingual models, such as mBERT, have shown promise in crosslingual transfer. In this work, we employ pruning to quantify the robustness and interpret layer-wise importance of mBERT. 
On four GLUE tasks, the relative drops in accuracy due to pruning have almost identical results on mBERT and BERT suggesting that the reduced \textit{attention capacity} of the multilingual models does not affect robustness to pruning.
For the crosslingual task XNLI, we report higher drops in accuracy with pruning indicating lower robustness in crosslingual transfer. 
Also, the importance of the encoder layers sensitively depends on the language family and the pre-training corpus size. 
The top layers, which are relatively more influenced by fine-tuning, encode important information for languages similar to English (SVO) while the bottom layers, which are relatively less influenced by fine-tuning, are particularly important for agglutinative and low-resource languages.
\end{abstract}

\begin{figure*}
    \begin{tabular}{@{\hskip 0in}cc}
        \includegraphics[width=4.45in]{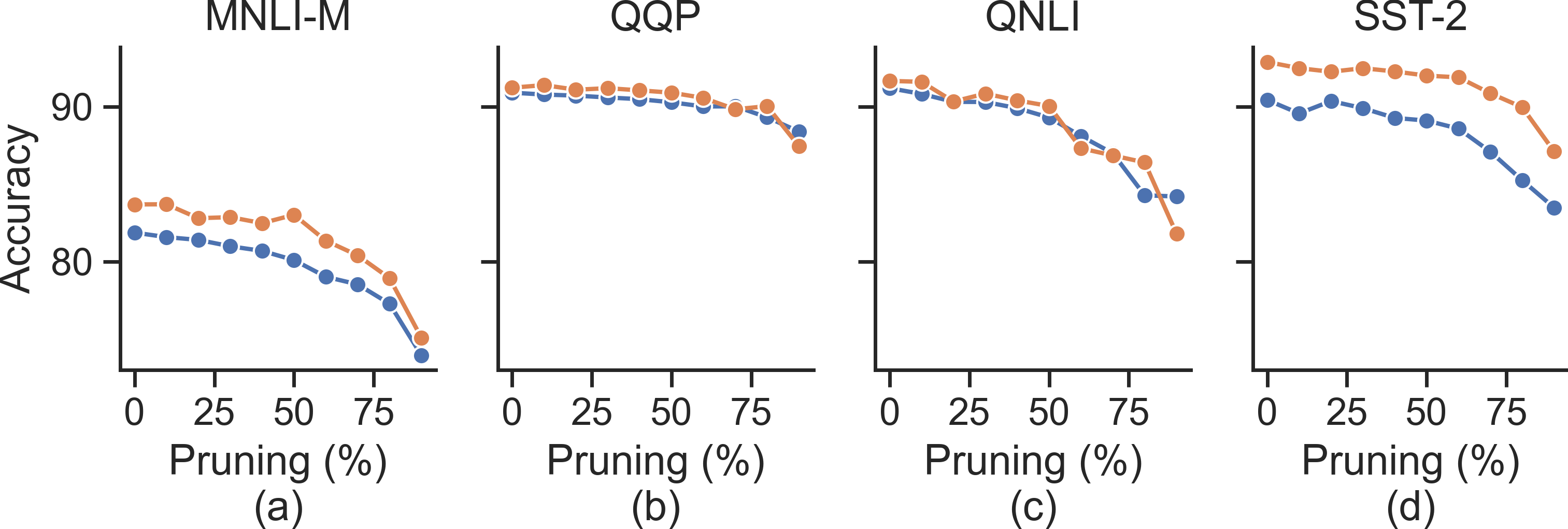} &  \includegraphics[width=1.6in]{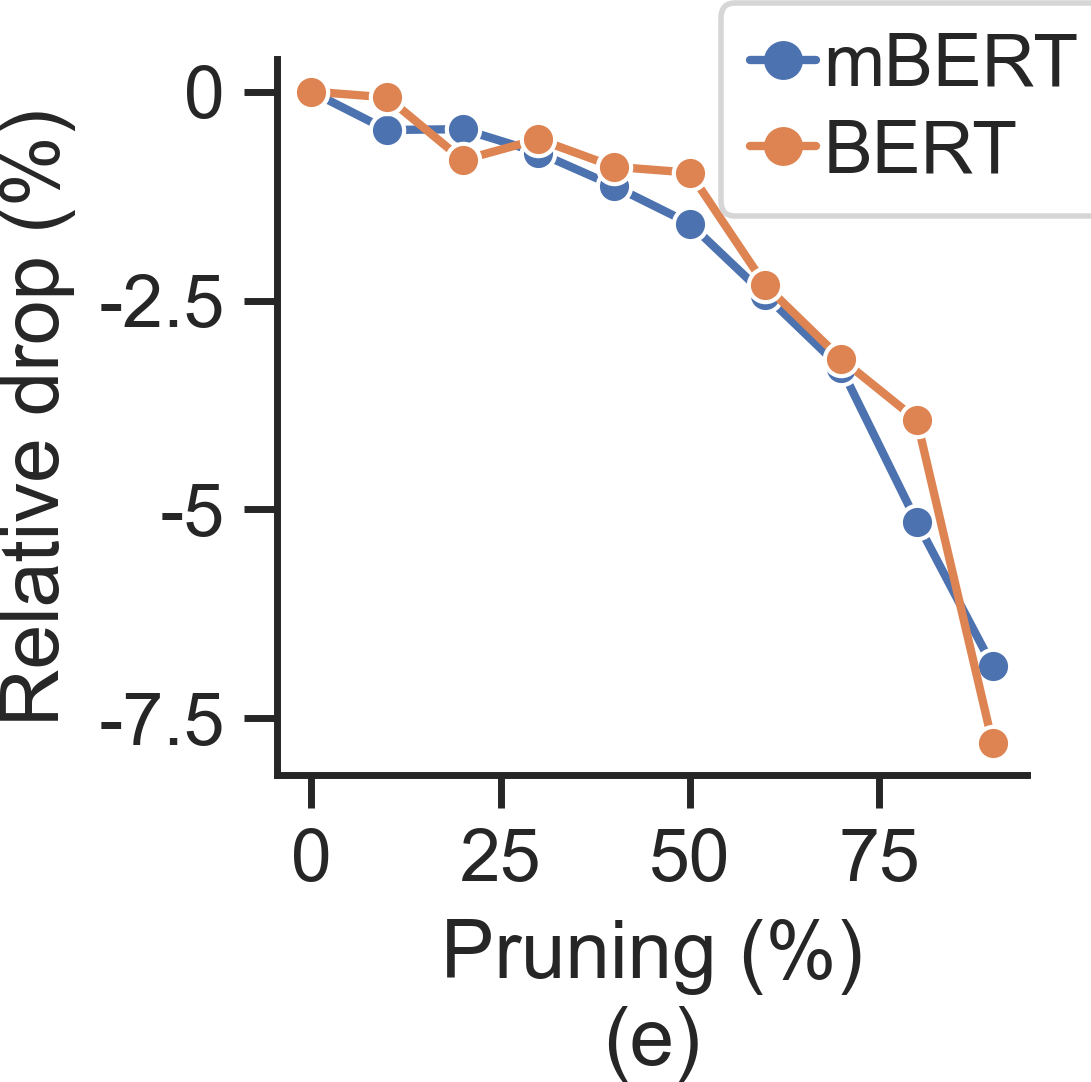} \\
    \end{tabular}
    \caption{Random Pruning - Effect of pruning attention heads in BERT and mBERT on 4 GLUE tasks [(a) to (d)]; and relative percentage drop in accuracy averaged across the four tasks (e).}
    \label{fig:random_pruning}
\end{figure*}

\section{Introduction}

Transformer-based models continue to achieve state-of-the-art performance on a number of NLU and NLG tasks. 
A rich body of literature, termed BERTology \cite{Rogers2020API}, has evolved to analyze and optimize these models. 
One set of studies comment on the functional role and importance of attention heads in these models \cite{clark2019does,michel2019sixteen, voita2019analyzing, voita2019bottom,liu2019attentive,belinkov2017neural}.
Another set of studies have identified ways to make these models more efficient by methods such as pruning \cite{mccarley2019pruning, gordon2020compressing, sajjad2020poor, budhraja2020weak}.
A third set of studies show that multilingual extensions of these models, such as Multilingual BERT \cite{devlin-etal-2019-bert}, have surprisingly high crosslingual transfer \cite{pires2019multilingual,wu-dredze-2019-beto}.%

Our work lies in the intersection of these three sets of methods: 
We analyze the importance of attention heads in multilingual models based on the effect of pruning on performance for both in-language and cross-language tasks.
We base our analysis on BERT and mBERT \cite{devlin-etal-2019-bert} and evaluate (i) in-language performance in English on four tasks from the GLUE benchmark - MNLI, QQP, QNLI, SST-2, and (ii) cross-language performance on the XNLI task on 10 languages - Spanish, German, Vietnamese, Chinese, Hindi, Greek, Urdu, Arabic, Turkish, and Swahili. With these, we derive two broad sets of findings.
Notice that we prune only the attention heads and not other parts of the network, such as the fully connected layers or the embedding layers. 
Thus all our results are restricted to comments on role of attention capacity in multilingual models. 

First, we compare and contrast the effect of pruning on in-language performance of BERT and mBERT. 
Intuitively, the reduced dedicated attention capacity in mBERT for English may suggest a more adverse effect of pruning on the GLUE tasks.
However, we find that mBERT is just as robust to pruning as is BERT.
At 50\% random pruning, average accuracy drop with mBERT on the GLUE tasks is 2\% relative to the base performance, similar to results for BERT reported in \citeauthor{budhraja2020weak}\shortcite{budhraja2020weak}.
Further, mBERT has identical preferences amongst layers to BERT, where (i) heads in the middle layers are more important than the ones in top and bottom layers, and (ii) consecutive layers cannot be simultaneously pruned.

Second, we study the effect of pruning on cross-language performance across languages that are categorised by language family (SVO/SOV/VSO/Agglutinative) and corpus size used for pre-training (high/medium/low-resource).
We find that mBERT is significantly less robust to pruning on the XNLI task: At 50\% random pruning, performance averaged across languages drops by 5\%.
However, this drop is not uniform across languages: The drop in SVO and high-resource languages is lower confirming that more pre-training data and similar language family %
make crosslingual transfer more robust.

Next, layer-wise pruning results reveal several insights into the functional roles of different layers in multilingual models.
Pruning bottom layers sensitively affects cross-language performance where the drop across languages shows a trend based on the language family: Agglutinative $>$ VSO $>$ SOV $>$ SVO.
On the other hand, pruning top layers has a lower impact on accuracy, but the order across language families is the exact reverse of the order for the bottom layers: SVO $>$ SOV $>$ VSO $>$ Agglutinative.
These results suggest that the bottom layers of the network are storing crosslingual information which is particularly crucial for performance on low-resource languages and languages quite different from the fine-tuned language (En-SVO). 
Further, the top layers of the network are specialising for the fine-tuned task and relatively more important for languages that are related to the fine-tuned language. 
We confirm this intuition by showing that the impact of fine-tuning on attention heads (captured by the difference in entropy of attention distributions across tokens) is much higher on the top layers than the middle or bottom layers.
We also observe that fine-tuning for a single epoch recovers about 93\% of cross-language performance consistently across all languages.

\begin{figure*}
    \includegraphics[width=6in]{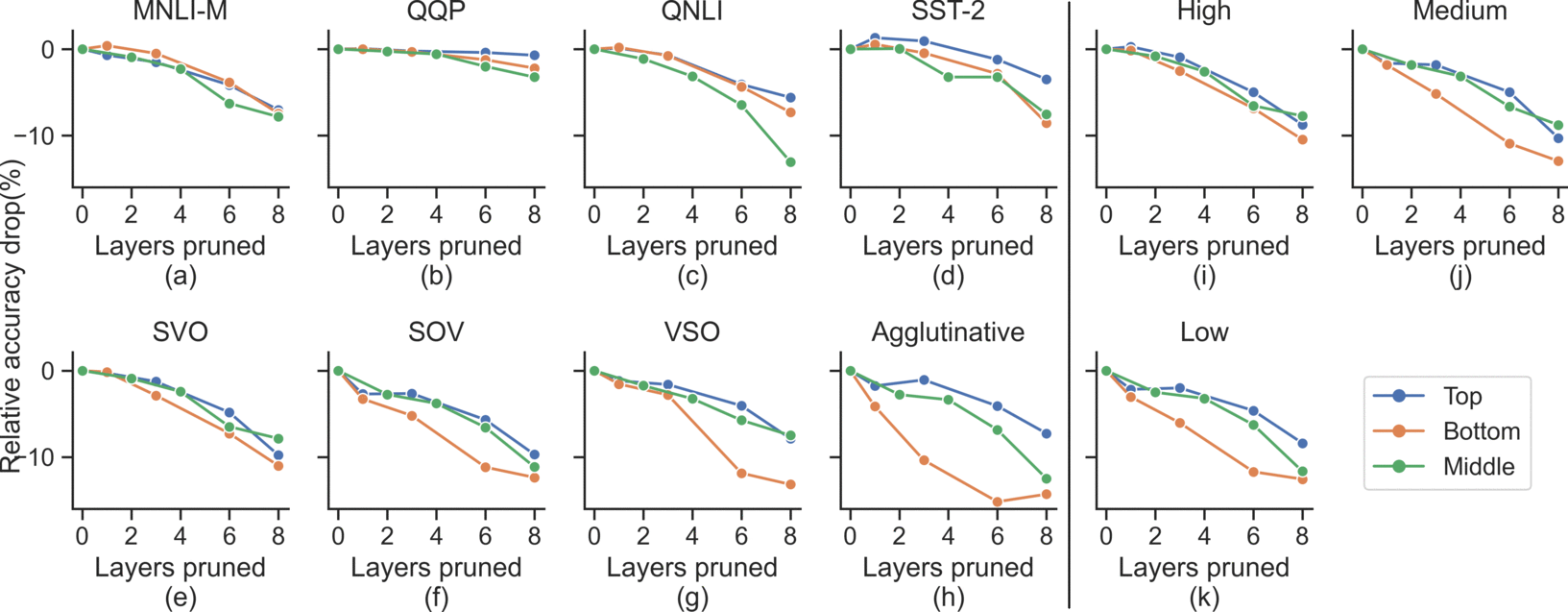}
    \caption{Layerwise pruning - Comparison across four GLUE tasks[(a)-(d)], and XNLI task: language family[(e)-(h)], and pre-training corpus size[(i)-(k)]}
    
    \label{label_combined_layers}
\end{figure*}

\section{In-language performance}

\citeauthor{DBLP:conf/acl/ConneauKGCWGGOZ20}\shortcite{DBLP:conf/acl/ConneauKGCWGGOZ20} argue that in multilingual models such as mBERT there is always a trade-off between \textit{transfer} and \textit{capacity dilution}. Specific to our focus on attention heads, the finite capacity of a fixed number of attention heads is shared by many languages. We refer to this as the \textit{attention capacity} of the network. As a result of this shared capacity, the in-language performance of multilingual models is typically poor when compared to monolingual models. Thus, the first question to ask is \textit{"Does pruning attention capacity affect the in-language performance of mBERT more adversely than BERT?"} Intuitively, if the attention capacity of mBERT is limited to begin with, then pruning it further should lead to large drops in performance. To check this intuition, we take the publicly available mBERT$_{BASE}$ model
which is pretrained on 104 languages.
We then fine-tune and evaluate it on four GLUE tasks: MNLI-M, QQP, QNLI, SST-2 \cite{wang2018glue}; and compare it with BERT$_{BASE}$. Following the evaluation setup as in \citeauthor{budhraja2020weak}\shortcite{budhraja2020weak}, we either (i) randomly prune k\% of attention heads in mBERT, or (ii) prune all heads in specific layers (top, middle, or bottom). In each case, after pruning we finetune the model for 10 epochs. Since the pruned attention heads are randomly sampled, we report the average across three different experiment runs. The standard deviation across three runs of experiments averaged across all languages is 0.51\% of the reported mean values. Our main observations are:%

\noindent \textbf{mBERT is as robust as BERT.} From Figure \ref{fig:random_pruning} we observe that at 0\% pruning, the performance of mBERT is comparable to BERT on 2 out of the 4 tasks (except MNLI-M and SST-2). Further, on pruning, the performance of mBERT does not drop drastically when compared to BERT. At all levels of pruning (0 to 90\% heads pruned) the gap between mBERT and BERT is more or less the same as that at the starting point (0\% pruning). This trend is more clear from Figure \ref{fig:random_pruning} (e) which shows the average drop in the performance at different levels of pruning relative to the base performance (no pruning). Even at 50\% pruning the average performance of mBERT drops by only 2\% relative to the base performance. Thus, contrary to expectation, mBERT is not adversely affected by pruning despite its seemingly limited attention capacity for each language.\\
\noindent \textbf{mBERT has same layer preferences as BERT.} From their pruning experiments, \citeauthor{budhraja2020weak}\shortcite{budhraja2020weak} show that the middle layers in BERT are more important than the top or bottom layers: The performance of BERT does not drop much when pruning top or bottom layers as compared to middle layers. We perform similar experiments with mBERT and find that the results are consistent. As shown in Figure \ref{label_combined_layers}) (a) to (d) across the 4 tasks mBERT does not have any preference amongst top and bottom layers, but middle layers are more important. This is especially true when we prune more layers ($>$4 out of the 12 layers in mBERT). \\
\noindent \textbf{mBERT doesn't prefer pruning consecutive layers (same as BERT).} Some works on BERTology \cite{lan2019albert} have identified that consecutive layers of BERT have similar functionality. We find supporting evidence for this in our experiments. In particular, pruning consecutive layers of mBERT as opposed to odd or even layers leads to a higher drop in the performance (see Table \ref{tab:mbert_glue}). %
Similar results are reported by \citeauthor{budhraja2020weak}\shortcite{budhraja2020weak} for BERT.
Thus, despite being multilingual, mBERT's layers have the same ordering of importance as BERT.

\begin{table}[tp]
\small
\begin{center}
\begin{tabular}{ c c c c c } 
 \hline
 \rule{0pt}{1.5ex} Layers Pruned & MNLI-M & QQP & QNLI & SST-2 \\
 \hline
 \rule{0pt}{1.5ex} Top six & 78.45 & 90.53 & 87.46 & 89.33 \\
 \rule{0pt}{1.5ex} Bottom six & 78.71 & 89.77 & 87.22 & 87.84 \\
 \rule{0pt}{1.5ex} Middle six & 76.72 & 89.07 & 85.28 & 87.5 \\
 \rule{0pt}{1.5ex} Odd six & 78.94 & 90.24 & 88.72 & 88.64 \\
 \rule{0pt}{1.5ex} Even six & 79.82 & 90.17 & 89.09 & 90.02 \\
 \hline
 
\end{tabular}
\caption{Effect of pruning self-attention heads of consecutive v/s non-consecutive layers of mBERT}
\label{tab:mbert_glue}
\end{center}
\end{table}

\section{Cross-language performance}

We now study the impact of pruning on the crosslingual performance of mBERT. Again, intuitively, one would expect that if we prune some heads then the crosslingual signals learned implicitly during training may get affected resulting in poor performance on downstream tasks. To check this intuition, we perform experiments with 11 languages on the XNLI dataset \cite{conneau2018xnli}. 
We categorize these languages according to their structural similarity as SVO (English, Spanish, German, Vietnamese, Chinese),
SOV (Hindi, Greek, Urdu),
VSO (Arabic),
and Agglutinative (Turkish, Swahili). Further, we follow \citeauthor{wu2020all}\shortcite{wu2020all} and classify these languages as High Resource (English, Spanish, German), Medium Resource (Arabic, Vietnamese, Chinese, Turkish), and Low Resource (Hindi, Greek, Urdu, Swahili) based on the size of the pretraining corpus. %
Similar to our in-language experiments reported above, we either (i) randomly prune k\% of attention  heads in mBERT, or (ii) prune all heads in specific layers (top, middle, or bottom). In each case, after pruning we finetune the model for 10 epochs on the task specific training data.  Our mains observations are: \\
\begin{figure}[tp]
    \includegraphics[width=2.7in]{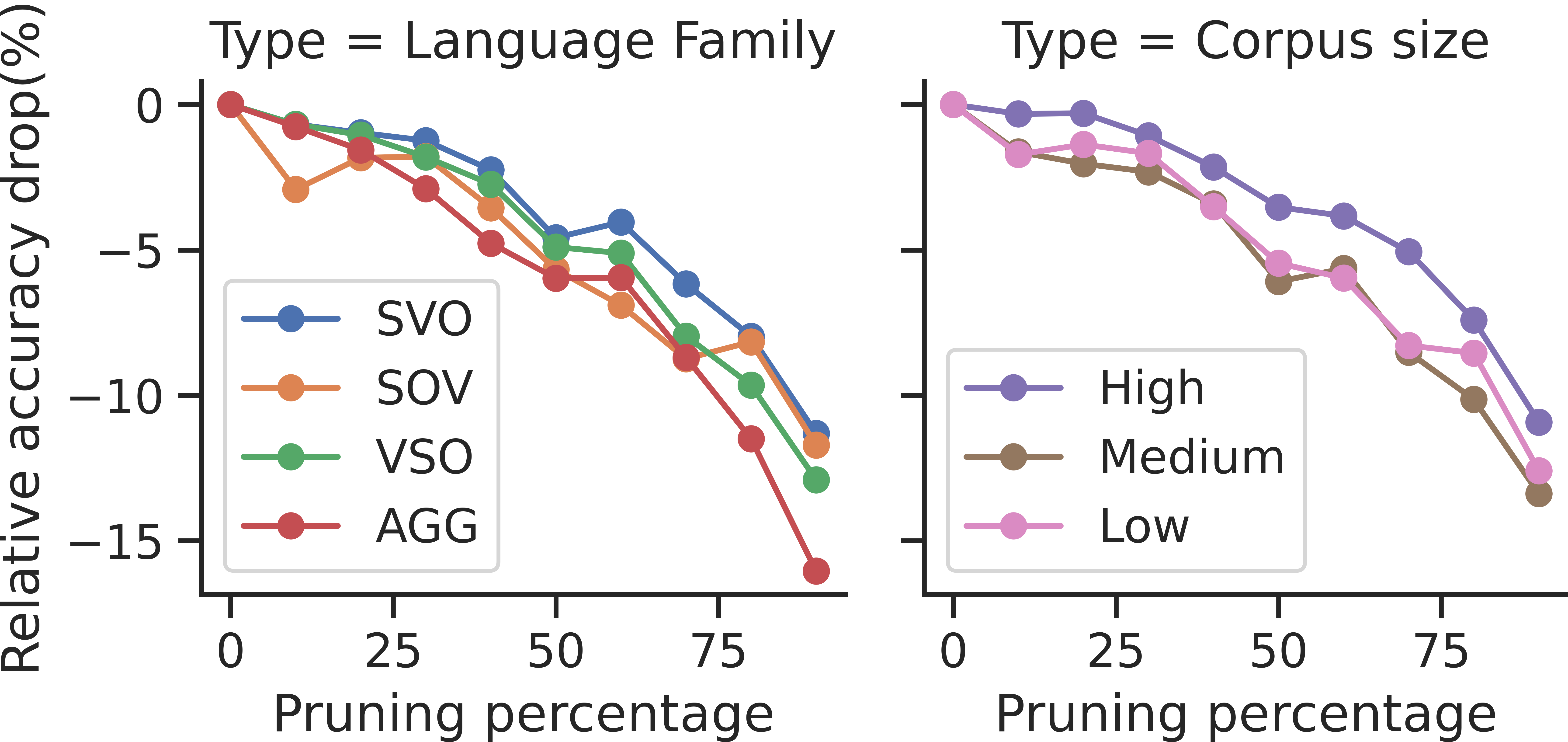}
    \caption{XNLI task - Performance of various language families at different percentages of pruning.}
    \label{label_xnli_pp}
\end{figure}
\noindent\textbf{mBERT is less robust in a cross language setup.} In Figure \ref{label_xnli_pp}, we plot the relative drop in performance from the baseline (0\% pruning) at different levels of pruning (10-90\%).
We observe that at 50\% pruning the relative drop in performance is around 5\% and at 90\% pruning the drop is around 10-15\%. We contrast this with the results in \ref{fig:random_pruning} (e) where the relative drop in performance for in-language tasks was more modest (around 2\% drop at 50\% pruning and 7.5\% drop at 90\% pruning). Thus, in a cross language setup mBERT is affected more adversely at higher levels of pruning (beyond 25\%). However, pruning upto 25\% of the heads does not affect the performance much ($\approx$ 2.5\% relative drop). 

\noindent\textbf{mBERT is more robust for SVO and high resource languages.} Referring to Figure \ref{label_xnli_pp}, we observe that high resource languages (more pretraining data) and SVO languages (similar to English) are relatively less adversely affected by pruning. Among the non-SVO languages,  agglutinative languages like Turkish are most affected by pruning\footnote{Note that, there is a high overlap between our SVO languages and high resource languages (since in the XNLI dataset most SVO languages are European languages which are high/mid resource languages)}.  

\noindent\textbf{mBERT is sensitive to pruning bottom layers for crosslingual transfer.}
Many recent works have shown that middle layers are important in BERT \cite{Pande2021TheHH, Rogers2020API, jawahar-etal-2019-bert, liu2019linguistic}. In this section, we analyse the relative importance of layers for different languages in a crosslingual task. Referring to Figure \ref{label_combined_layers} (e) to (k), we observe that for all languages, the bottom layers are more important than the top layers for crosslingual transfer. Specifically, we compare the difference $d(\cdot)$ between the relative performance drop when pruning the same number of top and bottom layers. We observe that $d(\mbox{SVO})$ $<$ $d(\mbox{SOV})$ $<$ $d(\mbox{VSO})$ $<$ $d(\mbox{Agglutinative})$ [(e) to (h) of Figure \ref{label_combined_layers}] and $d(\mbox{High~resource})$ $<$ $d(\mbox{Medium~resource})$  $<$ $d(\mbox{Low~resource})$ [(i) to (k) of Figure \ref{label_combined_layers}]. %
This suggests that most of the crosslingual information is stored in the bottom layers. While pruning top layers has a lower impact on accuracy, the order across language families is the exact reverse of the order for the bottom layers:  $d(\mbox{SVO})$ $>$ $d(\mbox{SOV})$ $>$ $d(\mbox{VSO})$ $>$ $d(\mbox{Agglutinative})$. Lastly, we observe that unlike BERT, mBERT is less sensitive to pruning middle layers for most languages.

\noindent\textbf{mBERT is more sensitive to pruning consecutive layers.}
We observe that mBERT is more sensitive to pruning consecutive layers. On average, pruning \textit{even} layers works better than pruning \textit{odd} or consecutive layers. This suggests that retaining the first layer (in even case) is beneficial.

\begin{figure}[tp]
\centering
  \begin{tabular}{@{\hskip 0in}c}
    \includegraphics[height=1.2in]{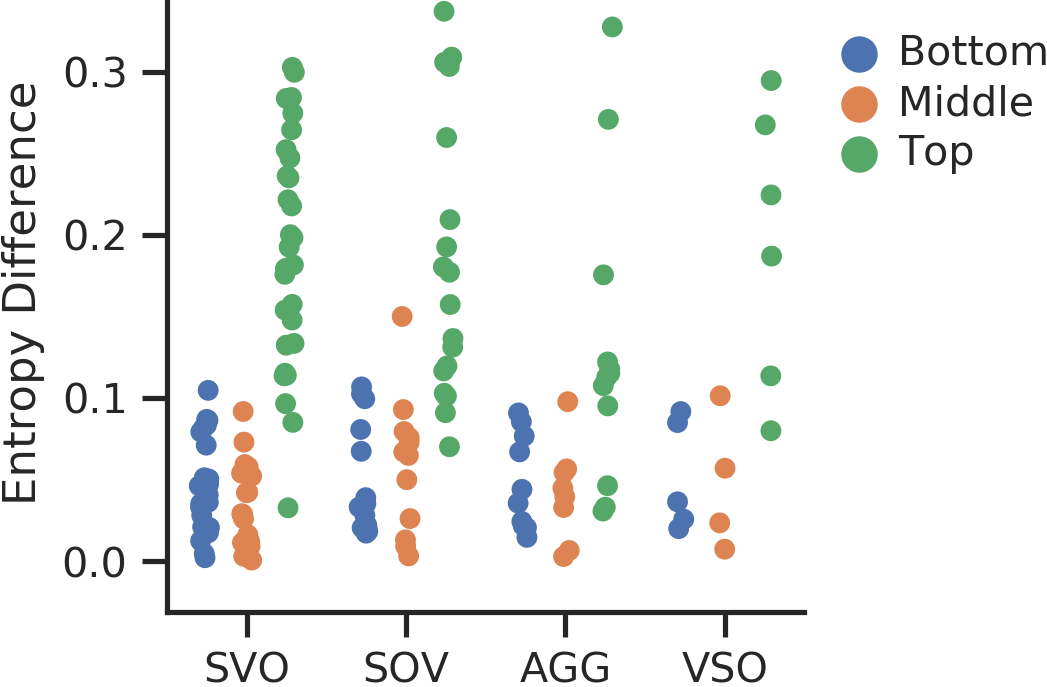} 
  \end{tabular}
    \caption{Change in entropy of alive heads before and after 1 epoch fine-tuning for 90\% rand. pruned model.}
    \label{label_entropy_change}
\end{figure}

\noindent\textbf{mBERT does not benefit from extensive fine-tuning after pruning.} We observe that on average (over all pruning experiments and languages on the XNLI task), mBERT achieves 93\% of the final accuracy after just one epoch of fine-tuning. Further fine-tuning causes only marginal gains in the accuracy. 
We then analyse the changes across layers to identify any preference for specific layers during fine tuning.
We then compute the entropy of each attention head in each layer for a given input token and average this entropy across all tokens in 500 sentences from the dev set. We plot the difference between the entropy of the unpruned heads before and after fine-tuning (see Figure \ref{label_entropy_change}). We notice that just like BERT \cite{kovaleva2019revealing}, the change in entropy is maximum for attention heads in the top layers (0.176) as compared to bottom (0.047) or middle layers (0.042). This  suggests that mBERT adjusts the top layers more during fine-tuning.

\section{Conclusion}
We studied the effect of pruning on mBERT and found that for in-language tasks, it is equally robust and has the same layer preferences as BERT.
For cross language tasks, mBERT is less robust with significant drops especially for low resource and non-SVO languages. Bottom layers are more important and pruning consecutive layers is not preferred. 
The importance of top and bottom layers have the reverse order across language families.

\bibliography{main}
\bibliographystyle{acl_natbib}

\appendix
\section{Appendix}
\label{sec:appendix}
\subsection{Model, dataset and hyperparameter details}
We use the publicly available multilingual BERT$_{BASE}$ model\footnote{https://git.io/Jt0A4 (by Google)} \cite{devlin-etal-2019-bert} which is pretrained on 104 languages. We fine-tune and evaluate this model by running each experiment on a single Google Cloud TPU (v2-8). We use the official development sets of the four tasks of the GLUE benchmark \cite{wang2018glue}: MNLI-M, QQP, QNLI and SST-2. We chose the best hyperparameters by trying all combinations of learning rate and batch size from \{2,3,4,5\} X 10$^{-5}$ and \{32, 64, 128\} respectively. For evaluation of the mBERT model in the crosslingual setting, we use the model finetuned on MNLI-M (English) data and report accuracies on the official development set of the XNLI dataset.

\subsection{Analysing sentence embeddings obtained from pruned mBERT}

\begin{table}[H]
    \begin{tabular}{@{\hskip 0in}c@{\hskip 0.01in}c}
    \includegraphics[width=3.7cm, height=4.5cm]{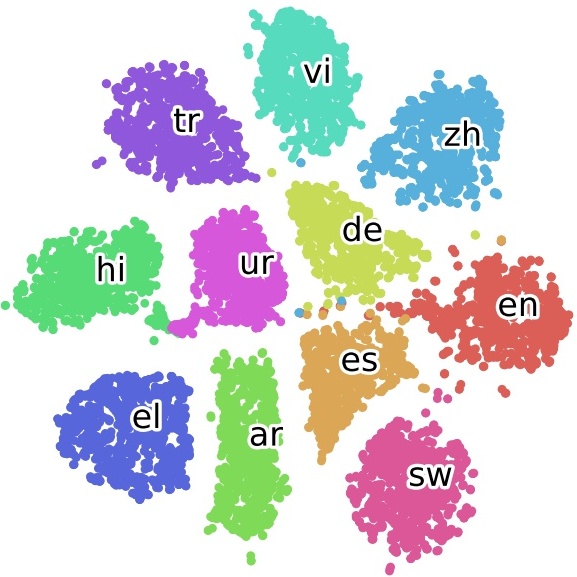} & \includegraphics[width=3.7cm, height=4.5cm]{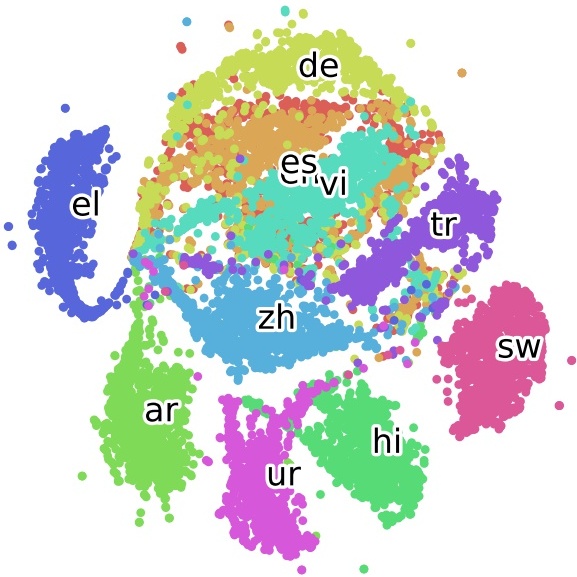} \\
    (a) & (b)\\
    \end{tabular}
    \captionof{figure}{t-SNE plots of sentence embeddings outputs of a 50\% randomly pruned model (a) before and (b) after one epoch of fine-tuning.}
    \label{label_tsne}
\end{table}

We take 1000 sentences per language and visualize the output embeddings of the sentences through t-SNE. Each of the 1000 sentences are given as input to a fifty percent pruned mBERT model. We observe the sentence embeddings at two stages: (i) Before fine-tuning and, (ii) After 1 epoch of fine-tuning.
We observe that before fine-tuning, we can clearly see separate clusters of all languages, whereas after one epoch of fine-tuning, we see that sentence embeddings of SVO languages (en, es, de, vi) are very close to each other, tend to overlap and hence become indistinguishable. Embeddings of SOV languages (hi and ur) also stay close, but the agglutinative languages (sw and tr) and VSO language (ar) don't seem to mix up well with the other languages.

\end{document}